\documentclass{article}

\usepackage[preprint]{neurips_2025}

\usepackage{microtype}
\usepackage{graphicx}
\usepackage{subfigure}
\usepackage{amsmath}
\usepackage{amsfonts}
\usepackage{booktabs}  
\usepackage{multirow}  
\usepackage{array}     
\usepackage{xcolor}    
\usepackage{hyperref}  
\usepackage{algorithm}
\usepackage{algpseudocode}

\title{SpecAttn: Speculating Sparse Attention}

\author{%
  Harsh Shah \\
  Machine Learning Department\\
  Carnegie Mellon University\\
  \texttt{hshah2@cs.cmu.edu} \\ 
}

\begin{document}

\maketitle

\begin{abstract}
Large Language Models (LLMs) face significant computational bottlenecks during inference due to the quadratic complexity of self-attention mechanisms, particularly as context lengths increase. We introduce SpecAttn, a novel training-free approach that seamlessly integrates with existing speculative decoding techniques to enable efficient sparse attention in pre-trained transformers. Our key insight is to exploit the attention weights already computed by the draft model during speculative decoding to identify important tokens for the target model, eliminating redundant computation while maintaining output quality. SpecAttn employs three core techniques: KL divergence-based layer alignment between draft and target models, a GPU-optimized sorting-free algorithm for top-p token selection from draft attention patterns, and dynamic key-value cache pruning guided by these predictions. By leveraging the computational work already performed in standard speculative decoding pipelines, SpecAttn achieves over 75\% reduction in key-value cache accesses with a mere  15.29\% increase in perplexity on the PG-19 dataset, significantly outperforming existing sparse attention methods. Our approach demonstrates that speculative execution can be enhanced to provide approximate verification without significant performance degradation.
\end{abstract}

\section{Introduction}
\label{introduction}

Transformer architectures have become the de-facto standard for sequence modeling across natural language processing tasks, owing to their ability to capture long-range dependencies via self-attention\cite{vaswani2023attentionneed}. However, the self-attention mechanism incurs time and memory costs that scale quadratically with sequence length, severely constraining model throughput and context window sizes in practice. To alleviate these inference bottlenecks, system-level engines such as vLLM employ optimized batching, GPU utilization, and key–value cache management to deliver 1.8–2.7× speedups over standard frameworks, yet they continue to execute full dense attention over the entire context \cite{kwon2023efficient}.

Complementary algorithmic solutions introduce sparse attention patterns to reduce pairwise token interactions. The Longformer work \cite{beltagy2020longformerlongdocumenttransformer} adopts sliding-window and global attention spans to achieve linear complexity with respect to sequence length 
while BigBird \cite{zaheer2021bigbirdtransformerslonger} combines local, random, and global attention heads to extend effective context lengths by up to eight times, with theoretical guarantees of universal approximation.

Despite their efficiency gains, these sparse architectures depend on predetermined attention patterns that are agnostic to individual input content and require model retraining or fine-tuning to adapt to new domains, limiting their flexibility for dynamic inference scenarios.

An alternative paradigm, speculative decoding, uses a lightweight draft model to generate candidate token sequences in parallel, which a larger verifier model then validates \cite{leviathan2023fastinferencetransformersspeculative}. This draft-and-verify approach achieves throughput improvements on large transformer models without altering output distributions or requiring weight updates.  

While prior research has treated speculative decoding and sparse attention as orthogonal optimization strategies, the attention patterns computed by draft models during speculative execution contain rich information about token importance that can be leveraged for dynamic, content-aware key-value cache pruning without much additional computational overhead.

We summarize our contributions below:
\begin{itemize}
    \item We provide a method to map draft model's layers to verifier model layers so that sparse attention can be speculated, using KL divergence as the similarity metric between attention distributions.
    
    \item We introduce a Triton kernel that identifies top-p nucleus tokens from the draft model's attention distribution without requiring sorting operations, enabling efficient GPU utilization by eliminating the computational overhead of explicit sorting.

\end{itemize}

\section{Related Works}

\textbf{System-Level Optimization.} Recent research has developed numerous techniques to accelerate LLM inference without sacrificing much accuracy. Microsoft's DeepSpeed library \cite{deepspeed} introduces kernel fusion and customized GPU kernels to speed up inference, achieving significant speedups on models like GPT-2 and BERT compared to PyTorch baseline inference. However, kernel fusion still computes the full attention matrix, making its latency grow steeply with sequence length. High-throughput KV-cache management in frameworks such as vLLM \cite{kwon2023efficient} and optimized servers like HuggingFace's Text Generation Inference harness smart batching and prefix caching to maximize GPU utilization and reuse computed states. FlashAttention \cite{flashattention1, flashattention2, flashattention3} leverages on-chip tiling and softmax fusion to reduce memory traffic and yields substantial latency reductions on long sequences. These methods deliver significant speedups on popular LLMs with minimal changes to existing models, but they still execute the full $O(n^2)$ attention: every query–key pair is computed regardless of its eventual impact on output quality.

\textbf{Sparse Attention Architectures.} Parallel to system-level advances, sparse-attention architectures and dynamic, content-dependent pruning approaches have aimed to break the quadratic barrier by only attending to a subset of tokens. Static sparse models such as Longformer \cite{beltagy2020longformerlongdocumenttransformer} and BigBird \cite{zaheer2021bigbirdtransformerslonger} impose hand-crafted patterns—sliding windows, global tokens, or random blocks—to achieve linear or near-linear complexity, but require retraining with those fixed masks and cannot adapt to the unique needs of each input. More recent inference-time methods such as MInference \cite{jiang2024minference10acceleratingprefilling} and SpargeAttn \cite{zhang2025spargeattnaccuratesparseattention} select top-k keys or apply two-stage filtering to drop low-importance attention links on the fly, offering faster prompt processing without retraining. Additional approaches include Quest \cite{tang2024quest}, which uses query-aware KV cache page selection, StreamingLLM \cite{xiao2023streamingllm}, which leverages attention sinks for infinite sequence length generalization, and Twilight \cite{lin2025twilightadaptiveattentionsparsity}, which applies adaptive top-p pruning for hierarchical attention sparsity. However, these semi-dynamic schemes still rely on predetermined head patterns or incur extra prediction overhead, and their benefits diminish on shorter contexts or during the token-by-token generation phase.

\textbf{Speculative Decoding.} Speculative decoding \cite{leviathan2023fastinferencetransformersspeculative, chen2023acceleratinglargelanguagemodel} offers an orthogonal speedup by pairing a small "draft" model with the full-scale LLM: the draft proposes tokens in batches that the verifier then checks, reducing the number of verifier model invocations. While this approach avoids retraining the main model, it does not change its internal attention cost and depends heavily on the draft's accuracy. Previous works have addressed speculative decoding and sparse attention as separate acceleration techniques, yet the attention distributions inherent to draft model computation provide a natural signal for identifying salient tokens. SpecAttn is motivated by these gaps: it unites speculative decoding's training-free paradigm with fully dynamic, content-aware sparse attention. By exploiting the attention weights already computed by the draft model to score and select the most informative tokens at each generation step, SpecAttn prunes the KV-cache significantly without any model modifications, preserving output fidelity while delivering end-to-end latency improvements.

\section{Method}

In this section, we present the methodology behind SpecAttn, our novel approach for accelerating inference in LLMs through dynamic sparse attention. We first introduce the problem formulation and then describe our framework in detail.

\subsection{Problem Formulation}
\label{subsec:problem_formulation}

\subsubsection{Sparse Attention Formulation}

Let $M_v$ denote the verifier model with $m$ layers and $M_d$ denote the draft model with $n$ layers. During the decoding phase of language model inference, we have the query vector $Q \in \mathbb{R}^{1 \times d}$ and the key-value cache $K, V \in \mathbb{R}^{L \times d}$, where $d$ is the head dimension and $L$ is the context length.

In standard dense attention, the output is computed as:

\begin{equation}
O = \text{softmax}\left(\frac{QK^T}{\sqrt{d}}\right)V = WV
\end{equation}

where $W = \text{softmax}\left(\frac{QK^T}{\sqrt{d}}\right) \in \mathbb{R}^{1 \times L}$ represents the normalized attention weights.

The computational complexity of this operation is $O(L^2d)$, which becomes prohibitively expensive for long sequences. To reduce this complexity, we introduce sparse attention by selecting only a subset of tokens to attend to.

\subsubsection{Sparse Attention with Token Selection}

Let $\mathcal{I}$ be the set of selected token indices. We define the sparse attention operation as:

\begin{equation}
\hat{O} = \text{softmax}\left(\frac{QK^T}{\sqrt{d}}\right)\Lambda_{\mathcal{I}}V = W\Lambda_{\mathcal{I}}V
\end{equation}

where $\Lambda_{\mathcal{I}} \in \mathbb{R}^{L \times L}$ is a diagonal mask matrix defined as:

\begin{equation}
\Lambda_{\mathcal{I}}[i, j] = 
\begin{cases}
1 & \text{if } i = j \text{ and } i \in \mathcal{I} \\
0 & \text{otherwise}
\end{cases}
\end{equation}

The theoretical error bounds and optimization objectives for this formulation are detailed in Appendix~\ref{app:error_bounds}.

\subsection{SpecAttn Framework Overview}
\label{subsec:framework}

\begin{figure}
  \centering
    \includegraphics[width=\linewidth]{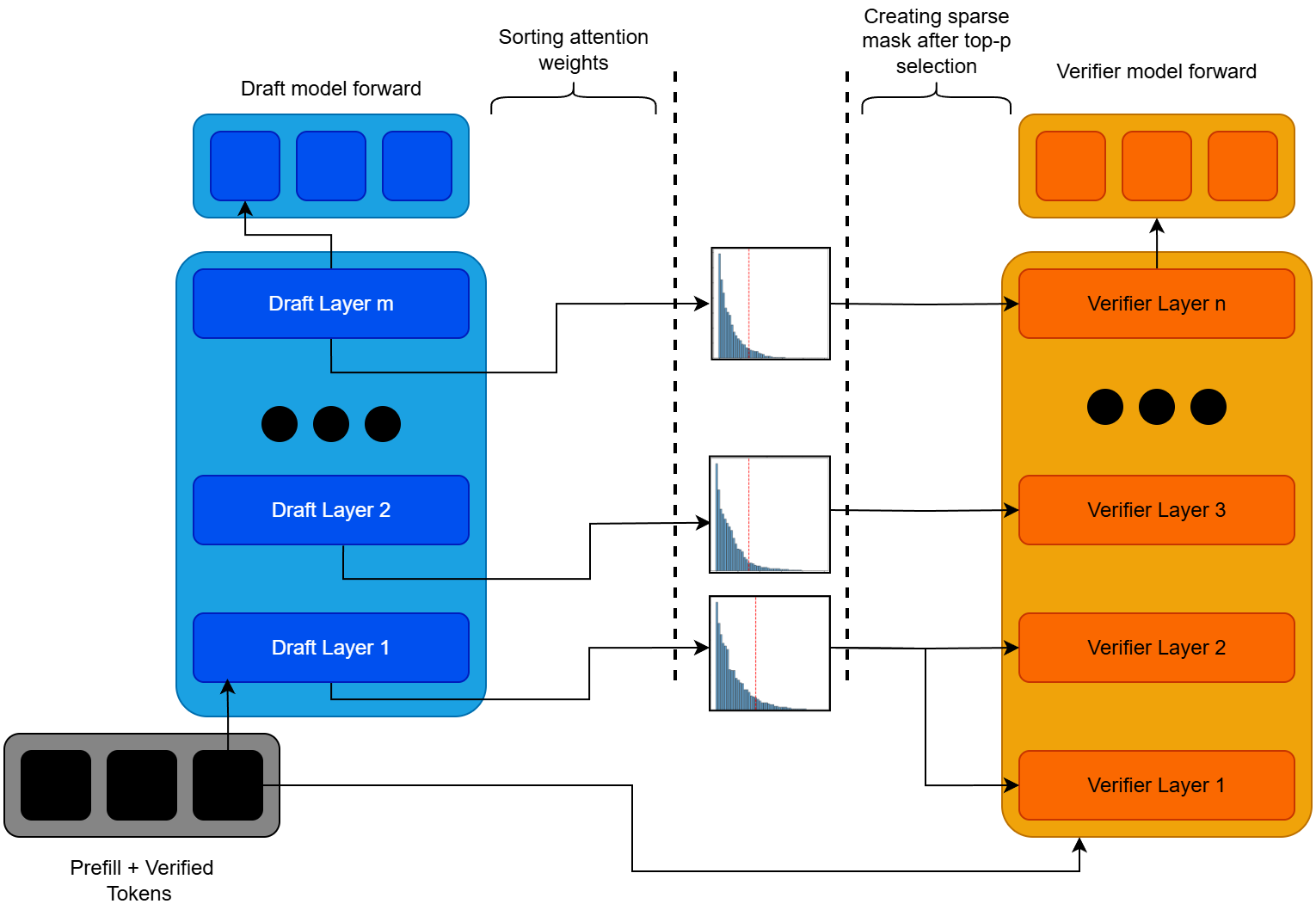}
  \caption{SpecAttn framework. Layer specific tokens are dynamically selected during runtime from draft model to prune KV cache of target model. Notice that a layer draft model can be mapped to multiple layers in verifier model. Also, the sorting described here is just for illustration, the implementation described in this paper performs the selection without sorting, Algo \ref{alg:sorting_free_nucleus}}
  \label{fig:architecture}
\end{figure}

Our SpecAttn framework leverages the complementary strengths of a lightweight draft model and a high-capacity verifier model in a cooperative inference pipeline. The core insight is that the attention patterns in the draft model often provide a good approximation of the important tokens that the verifier model should attend to. Figure~\ref{fig:architecture} illustrates the overall architecture of our approach.

The SpecAttn framework consists of three key steps:

\begin{enumerate}
    \item \textbf{Layer Mapping}: Establish correspondence between draft and verifier model layers using KL divergence similarity between attention distributions.
    \item \textbf{Token Selection}: Identify important tokens using sorting-free top-p nucleus selection from draft model attention patterns.
    \item \textbf{Sparse Attention Computation}: Perform sparse attention computation by attending only to selected tokens in the previous step.
\end{enumerate}

The overall workflow of SpecAttn is summarized in Algorithm \ref{alg:SpecAttn}

\begin{algorithm}
\label{alg:SpecAttn}
\caption{SpecAttn: Speculative Sparse Attention Framework}
\begin{algorithmic}[1]
\State \textbf{Input:} Input sequence $\mathbf{x} \in \mathbb{R}^{L}$, draft model $M_d$, verifier model $M_v$, lookahead $\gamma$, threshold $p$
\State \textbf{Output:} Generated sequence $\mathbf{y}$

\State \textbf{Initialization:}
\State Initialize KV caches: $\mathcal{K}_d, \mathcal{V}_d$ for draft, $\mathcal{K}_v, \mathcal{V}_v$ for verifier
\State Compute layer mapping $f: [m] \rightarrow [n]$ via $\arg\max_{i} \text{KL-sim}(A^v_j, A^d_i)$
\State Prefill both models: $M_d(\mathbf{x}) \rightarrow \mathcal{K}_d, \mathcal{V}_d$; $M_v(\mathbf{x}) \rightarrow \mathcal{K}_v, \mathcal{V}_v$
\State $\mathbf{y} \leftarrow \mathbf{x}$, $t \leftarrow |\mathbf{x}|$ \Comment{Initialize output and position counter}

\While{$t < T_{\max}$ and not EOS}
    \State \textbf{// Speculative Generation Phase}
    \State $\mathcal{S} \leftarrow \{\}$, $\mathcal{A} \leftarrow \{\}$ \Comment{Initialize buffers}
    
    \For{$s = 1$ to $\gamma$} \Comment{Generate $\gamma$ speculative tokens}
        \State $\mathbf{z}_s \leftarrow \mathbf{y}_{t-1}$ if \textit{s=1} \textbf{else} $\mathcal{S}_{s-1}$ \Comment{Use last accepted token}
        \State $\mathbf{o}^d_s, A^d_{1:n,s} \leftarrow M_d(\mathbf{z}_s, \text{cache}=\mathcal{K}_d, \mathcal{V}_d)$
        \State $\mathcal{S} \leftarrow \mathcal{S} \cup \{\arg\max(\mathbf{o}^d_s)\}$ \Comment{Store draft token}
        \State $\mathcal{A} \leftarrow \mathcal{A} \cup \{A^d_{1:n,s}\}$ \Comment{Store attention patterns}
        \State Update $\mathcal{K}_d, \mathcal{V}_d$ with $\mathbf{S_{-1}}$ \Comment{Update draft model's KV cache}
    \EndFor

    \State \textbf{// Sparse Attention Mask Creation}
    \For{$j = 1$ to $m$} \Comment{For each verifier layer}
        \State $i^* \leftarrow f(j)$ \Comment{Get corresponding draft layer index}
        \State $\mathcal{T}_j \leftarrow \text{SortingFreeNucleus}(\{ \mathcal{A}_{i^*}^d\}_{d=1}^{\gamma}, p)$ \Comment{Algorithm~\ref{alg:sorting_free_nucleus}}
        \State $\mathcal{M}_j \leftarrow \text{CreateSparseMask}(\mathcal{T}_j, |\mathcal{K}_v|)$ \Comment{Binary attention mask}
    \EndFor

    \State \textbf{// Verification with Sparse Attention}
    \State $\mathbf{H}^v, \mathbf{O}^v \leftarrow M_v(\mathcal{S}, \text{masks}=\{\mathcal{M}_j\}_{j=1}^m, \text{cache}=\mathcal{K}_v, \mathcal{V}_v)$
    \State $\hat{\mathbf{y}}_{1:\gamma+1} \leftarrow \arg\max(\mathbf{O}^v)$ \Comment{Verifier predictions}

    \State \textbf{// Token Verification \& Acceptance}
    \State $n_{acc} \leftarrow \text{CheckAcceptance}(\mathcal{S}_{1:\gamma}, \hat{\mathbf{y}}_{1:\gamma})$ \Comment{Count accepted tokens}
    \State $\mathbf{y}_{t+1:t+n_{acc}} \leftarrow \mathcal{S}_{1:n_{acc}}$ \Comment{Accept verified tokens}
    \State $t \leftarrow t + n_{acc} + 1$ \Comment{Update position counter}

    \State \textbf{// Cache Management}
    \State Update $\mathcal{K}_d, \mathcal{V}_d$ with accepted tokens $\mathbf{y}_{t-n_{acc}:t}$
    \State Update $\mathcal{K}_v, \mathcal{V}_v$ with accepted tokens $\mathbf{y}_{t-n_{acc}:t}$
\EndWhile

\State \Return $\mathbf{y}$
\end{algorithmic}
\end{algorithm}

\vspace{0.5cm}

\subsection{Layer Mapping via KL Divergence}
\label{subsec:layer_mapping}

A critical challenge in our approach is establishing an optimal mapping between the layers of the draft model and the verifier model, especially when these models have different depths. We formulate this as finding the layer in the draft model that has the most similar attention distribution to each layer in the verifier model. Note that this mapping is formed offline using a representative dataset (wikitext \cite{merity2016pointer}) and does not change during runtime.

Let $A^d_i \in \mathbb{R}^{L}$ and $A^v_j \in \mathbb{R}^{L}$ represent the attention distributions from layer $i$ of the draft model and layer $j$ of the verifier model, respectively. We define the similarity between these layers using the KL divergence:

\begin{equation}
S_{i,j} = -D_{KL}(A^v_j || A^d_i) = -\sum_{k=1}^{L} A^v_j[k] \log \frac{A^v_j[k]}{A^d_i[k]}
\end{equation}

For each verifier layer $j$, we find the best matching draft layer while keeping the index chosen for draft layer monotonically increasing. The intuition behind monotonic mapping is that the attention patterns are similar for different models as one moves up the models' layers. The mapping problem is a modification to dynamic time warping and can be efficiently solved by dynamic programming techniques. Pseudo code and more details regarding the mapping can be found in appendix \ref{app:layer_mapping_algo}. 

\subsection{Sorting-Free Top-p Nucleus Selection}
\label{subsec:token_selection}

Once we have established the layer mapping, we leverage the draft model's attention distributions to determine which tokens the verifier model should attend to. Traditional top-p selection requires sorting the attention weights, which can be computationally expensive on GPUs. We propose a sorting-free algorithm for efficient nucleus identification inspired from work in \cite{lin2025twilightadaptiveattentionsparsity}. 

\begin{algorithm}
\caption{Sorting-Free Top-p Nucleus Selection}
\label{alg:sorting_free_nucleus}
\begin{algorithmic}[1]
\Procedure{SortingFreeNucleus}{$\mathcal{A}_{layer} \in \{\mathbb{R}^{1 \times L}\}_{\gamma}$, $p \in (0,1]$}
    \State $\mathcal{T} \leftarrow \{\}$ \Comment{Initialize combined token set}
    
    \For{$s = 1$ to $\gamma$} \Comment{Process each speculative step}
        \State $\mathbf{a}_s \leftarrow \mathcal{A}_{layer,s}$ \Comment{Get attention weights for step $s$}
        \State $M_{target} \leftarrow p \cdot \sum_{i=1}^{L} a_{s,i}$ \Comment{Target attention mass}
        \State $\theta_{high} \leftarrow \max(\mathbf{a}_s)$, $\theta_{low} \leftarrow 0$ \Comment{Binary search bounds}
        \State $\epsilon \leftarrow 10^{-6}$ \Comment{Convergence tolerance}
        
        \While{$\theta_{high} - \theta_{low} > \epsilon$}
            \State $\theta_{mid} \leftarrow \frac{\theta_{high} + \theta_{low}}{2}$
            \State $M_{current} \leftarrow \sum_{i: a_{s,i} \geq \theta_{mid}} a_{s,i}$ \Comment{Mass above threshold}
            \If{$M_{current} < M_{target}$}
                \State $\theta_{high} \leftarrow \theta_{mid}$ \Comment{Lower threshold}
            \Else
                \State $\theta_{low} \leftarrow \theta_{mid}$ \Comment{Raise threshold}
            \EndIf
        \EndWhile
        
        \State $\mathcal{T}_s \leftarrow \{i : a_{s,i} \geq \theta_{mid}\}$ \Comment{Selected tokens for step $s$}
        \State $\mathcal{T} \leftarrow \mathcal{T} \cup \mathcal{T}_s$ \Comment{Union with global token set}
    \EndFor
    
    \State \Return $\mathcal{T}$ \Comment{Return combined token indices}
\EndProcedure
\end{algorithmic}
\end{algorithm}

This binary search approach finds the optimal threshold without explicit sorting, making it more GPU-friendly. For each layer $j$ in the verifier model, we:

\begin{enumerate}
    \item Extract attention weights $\mathbf{a}_{f(j)} \in \mathbb{R}^L$ from the corresponding draft model layer $f(j)$.
    \item Apply Algorithm~\ref{alg:sorting_free_nucleus} \footnote{the implementation in this work uses a fixed number of iterations (i.e., 10)} to find the minimal subset of tokens $\mathcal{T}_j \subseteq \{1, \ldots, L\}$ such that:

    $\sum_{k \in \mathcal{T}_j} a_{f(j),k} \geq p$
    where $p \in (0, 1]$ is a predefined threshold (e.g., $p = 0.95$).
\end{enumerate}

This approach corresponds to performing top-$p$ (nucleus) sampling over the attention weights, ensuring that we retain tokens accounting for at least $p$ fraction of the total attention mass. The parameter $p$ allows us to control the trade-off between computational efficiency and output quality.

\subsection{Sparse Attention Computation}
To compute sparse attention with the selected tokens, we employ the sparse attention kernel from Flashinfer \cite{ye2025flashinfer}. The generated mask containing selected tokens is converted to compressed sparse row (CSR) format, which introduces additional latency during generation, before being used for sparse attention computation.
\section{Experiments}
\label{sec:experiments}
\begin{figure}
  \centering
    \includegraphics[width=0.7\linewidth]{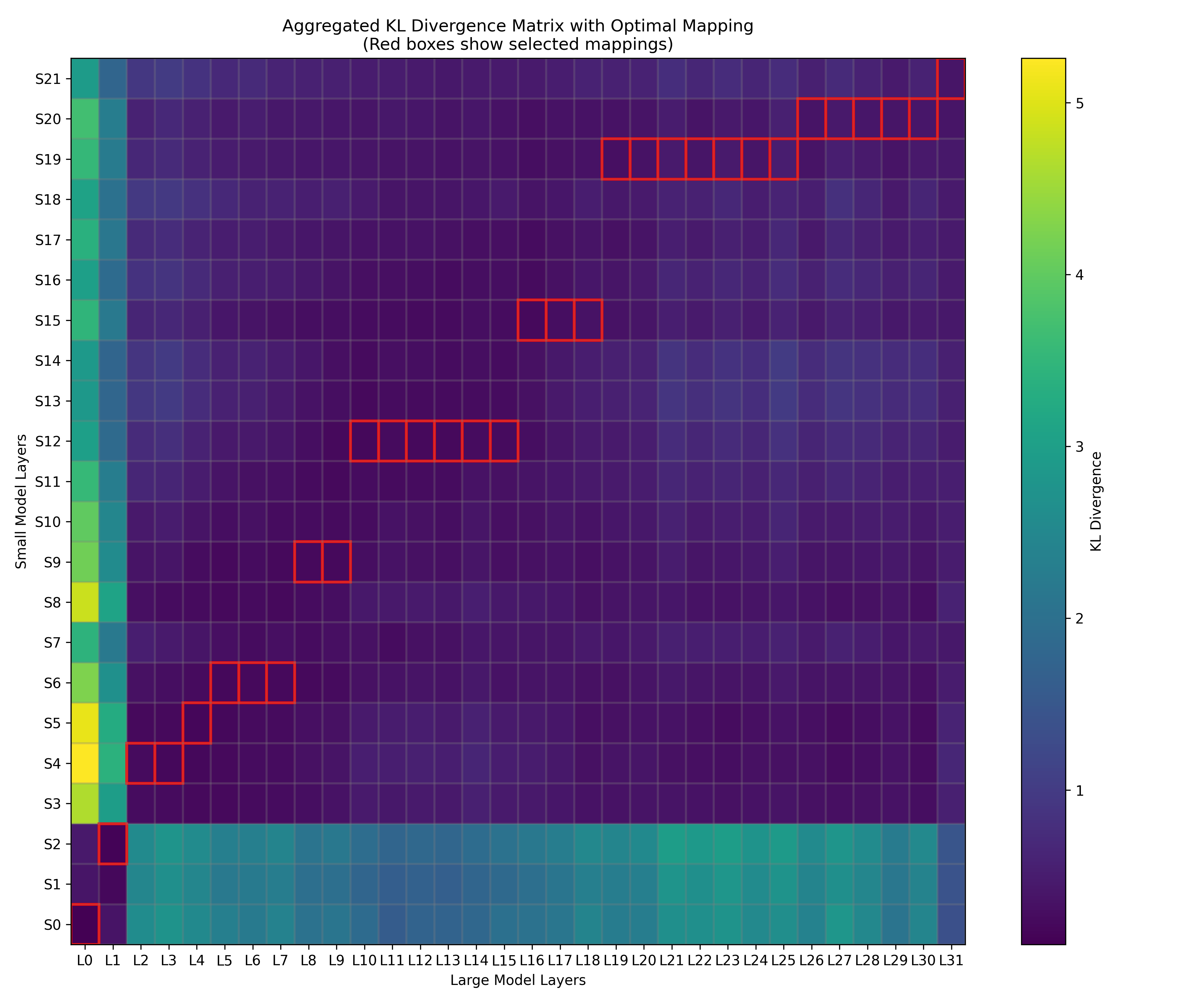}
  \caption{Heatmap of KL divergence between layers of \textbf{TinyLlama-1.1B} as small model (draft model) and \textbf{Llama-2-7b-hf} as large model (verifier model). Red boxes denote the draft layer selected for the corresponding verifier layer.}
  \label{fig:mapping}
\end{figure}

We have conducted all our experiments using a single NVIDIA RTX 4090 GPU, with 24GB VRAM. For all experiments in this section, we use \textbf{TinyLlama-1.1B}\footnote{https://huggingface.co/TinyLlama/TinyLlama-1.1B-intermediate-step-1431k-3T} as draft model and \textbf{Llama-2-7b-hf}\footnote{https://huggingface.co/meta-llama/Llama-2-7b-hf} as the target model. The mapping for these models obtained by method described in \ref{subsec:layer_mapping} is shown in Fig \ref{fig:mapping}. For analsyis of perplexity, we use PG-19 dataset \cite{raecompressive2019} truncated to 2048 tokens, out of which 10\% are used for prefill and other 90\% are used for evaluating decoding process, simulating long generation scenarios. We also measure KV computation reduction and end-to-end latency on LongBench \cite{bai2024longbench} task gov\_report to compare SpecAttn with full attention speculative decoding (with the similar prefill and decoding setup as for perplexity analysis).

\subsection{Perplexity Evaluation}

We evaluate the perplexity of different sparse attention methods to assess their impact on generation quality. Table~\ref{tab:perplexity_comparison} shows the comparison between our method and existing sparse attention techniques. Note that the first 2 layers use full attention for all methods due to diffused attention found in initial layers (described in \cite{tang2024quest}). Further the chunk size of SpecAttn is set to 16 for fair comparison with Quest (\cite{tang2024quest}). It can be observed that SpecAttn (p=0.95) reduces KV cache loading by 78.4\% while increasing perplexity by mere 0.984 (15.29\% relative increase), which significantly improves upon other sparse attention methods. 

\begin{table}[h]
\centering
\caption{Performance Comparison of Sparse Attention Methods}
\begin{tabular}{lcccc}
\toprule
Method & Perplexity & Perp. Diff. & Rel. Increase & KV Reduction \\
\midrule
\textbf{Full Attention} & 6.435 & - & - & - \\
\cmidrule[0.4pt](lr){1-5}
\textbf{StreamingLLM (\cite{xiao2023streamingllm}) } & 186.242 & +179.807 & +2794.32\% & 77.4\% \\
\textbf{Quest (\cite{tang2024quest})} & 7.823 & +1.389 & +21.58\% & 77.4\% \\
\textbf{SpecAttn (p=0.95)} & 7.419 & +0.984 & +15.29\% & \textbf{78.4}\% \\
\textbf{SpecAttn (p=0.97)} & 6.720 & +0.285 & +4.43\% & 68.8\% \\
\textbf{SpecAttn (p=0.99)} & \textbf{6.471} & \textbf{+0.036} & \textbf{+0.56\%} & 44.3\% \\
\bottomrule
\end{tabular}
\label{tab:perplexity_comparison}
\end{table}

\subsection{Speedup Analysis}
\begin{figure}
  \centering
  \begin{minipage}{0.49\textwidth}
    \centering
    \includegraphics[width=\linewidth]{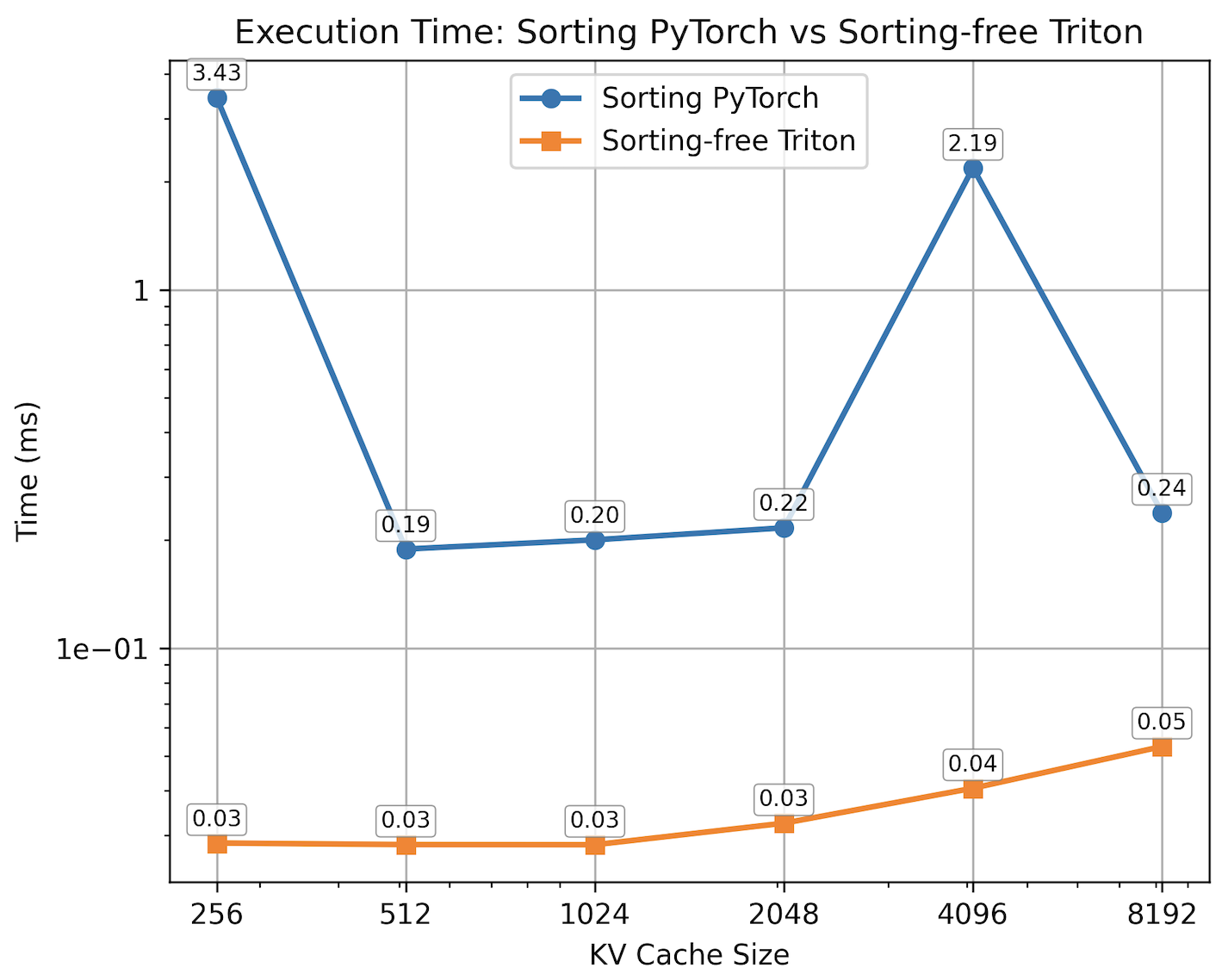}
    \caption{Graph of time taken for mask generation when comparing sorting in pytorch with sorting-free algorithm as depicted in Algo \ref{alg:sorting_free_nucleus}}
    \label{fig:kv_cache_size_vs_time}
  \end{minipage}%
  \hfill
  \begin{minipage}{0.49\textwidth}
    \centering
    \includegraphics[width=\linewidth]{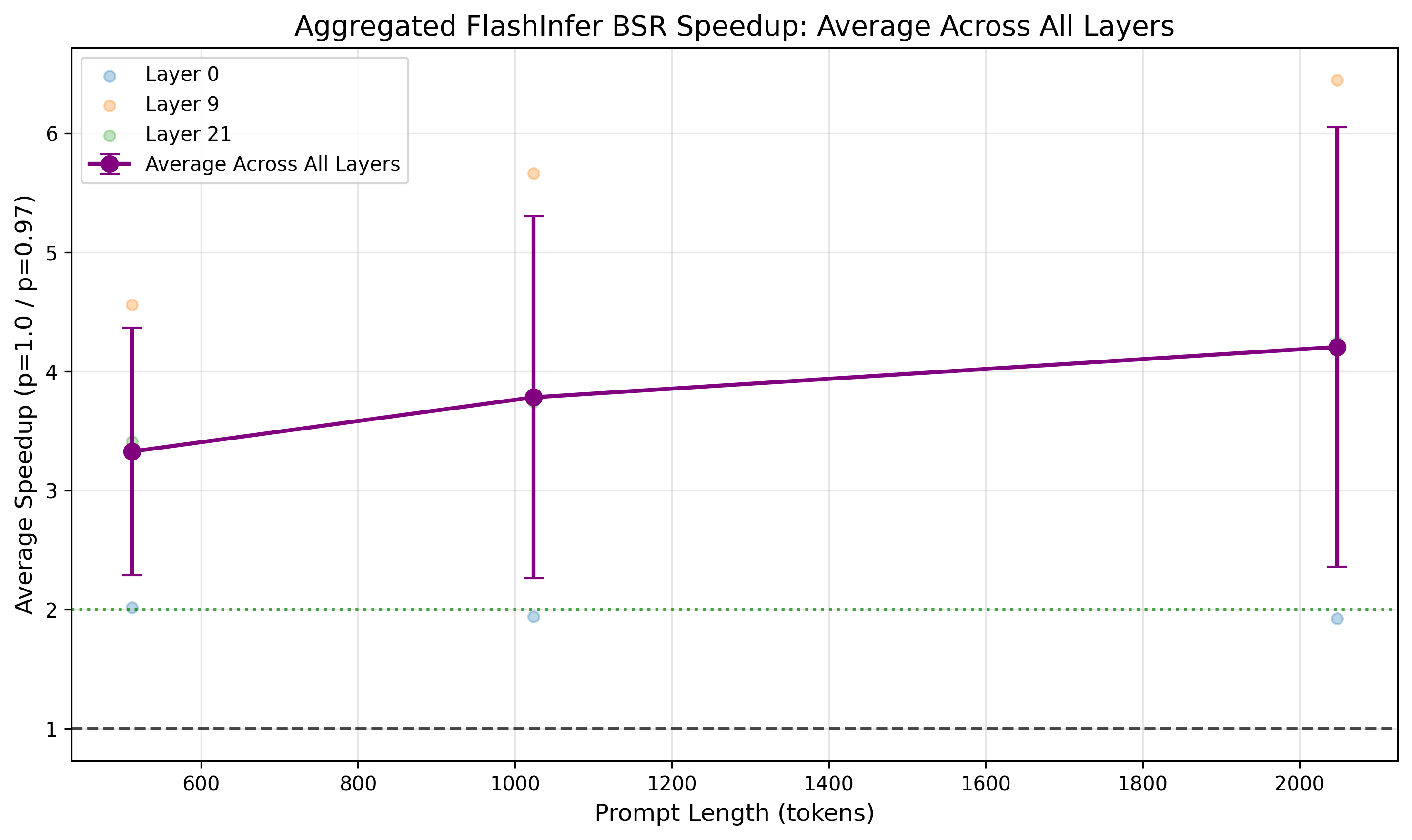}
    \caption{Speedup achieved in computing attention through FlashInfer's BlockSparseAttention with p=0.97 as compared to p=1.0 (i.e., full attention)}
    \label{fig:aggregated_speedup_vs_prompt_length}
  \end{minipage}
\end{figure}

We show the speedup achieved by using sorting-free triton kernel (as described in \ref{alg:sorting_free_nucleus}) when compared to sorting-based algorithm, in Fig \ref{fig:kv_cache_size_vs_time}. It can be observed that there is at least 4x speedup till KV cache size 8192 by uing sorting-free trtion kernel. For computing sparse attention, we use Flashinfer's \cite{ye2025flashinfer} BlockSparseAttention \footnote{https://docs.flashinfer.ai/api/sparse.html}. In Fig \ref{fig:aggregated_speedup_vs_prompt_length}, we show the speedup obtained when using p=0.97 as compared to p=1.0 (full attention). It shows an increasing trend of speedup as prompt length increases, with more than 4x speedup for prompt length 2048. \\
We also evaluated SpecAttn using end-to-end latency comparison with full attention speculative decoding (Table \ref{tab:method_comparison_total_time}). The end-to-end latency seems to be higher for SpecAttn, and this can be attributed to high mask generation time (Algo \ref{alg:sorting_free_nucleus}) which is compensated through sparsity in KV cache. However, the trend in Fig \ref{fig:aggregated_speedup_vs_prompt_length} gives promising direction of increasing the context length for the experiments, since that would lead to further reduction of latency in attention computation, compensating for the mask generation time.

\begin{table}[htbp]
  \centering
  \caption{Throughput experiments}
  \label{tab:latency_experiments}
  \begin{tabular}{lccc}
    \hline
    Method & Tokens/sec ($\uparrow$) & KV Reduction ($\uparrow$) \\
    \hline
    No Spec. Decoding (flashattn) & 42.00 & - \\
    Spec. Decoding (Full Attention) & \textbf{68.26} & - \\
    SpecAttn (p=0.97) & 59.95 & \textbf{71.89\%} \\
    \hline
  \end{tabular}
  \label{tab:method_comparison_total_time}
\end{table}

\section{Conclusion}
In this work, we introduced SpecAttn, a training-free framework that fuses speculative decoding with dynamic sparse attention to drastically reduce the key-value (KV) cache usage in LLM inference. By mapping layers using KL divergence similarity and employing a sorting-free top-p token selection mechanism at each decoding step, SpecAttn achieves significant KV cache reductions and maintains competitive perplexity compared to existing sparse attention methods.

Our empirical evaluations demonstrate that SpecAttn can reduce KV cache access by up to 78\% while maintaining reasonable perplexity degradation (15.29\% relative increase for p=0.95). The method excels at balancing efficiency and quality, outperforming existing methods like Quest at similar sparsity levels through context-aware and dynamic pruning.

\section{Limitations and Future Work}
\label{sec:limitations}
Several promising directions remain for future investigation. While we focused on KL divergence for layer mapping, exploring alternative similarity metrics (jaccard similarity, other distribution distances, etc.) could potentially improve the quality of layer correspondences between draft and verifier models. Additionally, extending our evaluation to much longer contexts (10K+ tokens) would provide better understanding of SpecAttn's scaling properties and effectiveness in truly long-context scenarios where the quadratic attention cost becomes more prohibitive. Finally, integrating SpecAttn into production serving frameworks like vLLM to leverage PagedAttention and emerging dynamic token caching (DTC) capabilities would enable comprehensive benchmarking and real-world deployment validation.


\appendix

\section{Error Bounds and Optimization Objective}
\label{app:error_bounds}

In this appendix, we provide the theoretical foundation for our sparse attention formulation introduced in the main paper. We establish error bounds for the sparse attention approximation and derive the optimization objective that motivates our top-p token selection strategy.

\subsection{Problem Context and Motivation}
Large language models compute attention over increasingly long sequences, leading to quadratic computational complexity $O(L^2d)$ where $L$ is the sequence length and $d$ is the head dimension. Our sparse attention framework addresses this by selecting only a subset of tokens to attend to, reducing both computational cost and memory requirements.

Given the standard dense attention output $O = \text{softmax}\left(\frac{QK^T}{\sqrt{d}}\right)V = WV$ and our sparse approximation $\hat{O} = W\Lambda_{\mathcal{I}}V$ where $\mathcal{I}$ represents selected token indices, we seek to minimize the approximation error while maximizing computational savings.

\subsection{Error Bound and Optimization Objective}
The output error can be bounded as follows:
\begin{align}
\mathcal{L} &= \|O - \hat{O}\| = \|W(\Lambda_{\mathcal{I}} - I_{L \times L})V\| \\
&\leq \|W(\Lambda_{\mathcal{I}} - I_{L \times L})\| \cdot \|V\|
\end{align}

Our objective becomes minimizing $\|W(\Lambda_{\mathcal{I}} - I_{L \times L})\| = 1 - \sum_{i \in \mathcal{I}} W[i]$. This means we want to select tokens that maximize the sum of attention weights, preserving the most important attention mass.

\subsection{From Top-k to Top-p Sparse Attention}
Traditional approaches to sparse attention use a fixed budget $B$ (top-k) to select tokens:
\begin{equation}
\mathcal{I} = \arg\max_{\mathcal{I}} \sum_{i=1}^{L} W\Lambda_{\mathcal{I}} \quad \text{s.t.} \quad |\mathcal{I}| = B
\end{equation}

However, a major limitation of the top-k approach is that it cannot adapt to different attention weight distributions. Attention patterns can vary significantly between flat (diffuse) and peaked (focused) distributions. A fixed budget $B$ would either be inefficient for peaked distributions or insufficient for flat distributions.

To address this issue, we propose using a top-p (nucleus) approach for token selection:
\begin{equation}
\mathcal{I} = \arg\min_{\mathcal{I}} |\mathcal{I}| \quad \text{s.t.} \quad \sum_{i \in \mathcal{I}} W[i] \geq p
\end{equation}

where $p \in (0, 1]$ is a threshold parameter. This formulation dynamically adjusts the number of tokens based on the attention distribution, selecting just enough tokens to capture $p$ fraction of the total attention mass.

The top-p approach provides a theoretical error bound of $(1-p) \cdot \|V\|$, making it possible to control the trade-off between computational efficiency and output quality by adjusting the parameter $p$.

\section{Layer Mapping Algorithm}
\label{app:layer_mapping_algo}
The layer mapping problem can be formulated as finding an optimal monotonic alignment between draft model layers and verifier model layers. Given a similarity matrix $S \in \mathbb{R}^{m \times n}$ where $S[i,j]$ represents the compatibility score between draft layer $i$ and verifier layer $j$, we seek a mapping function $f: \{1, 2, \ldots, n\} \rightarrow \{1, 2, \ldots, m\}$ that maximizes the total alignment score $\sum_{j=1}^{n} S[f(j), j]$ subject to monotonicity constraints. The monotonicity requirement ensures that $f(j_1) \leq f(j_2)$ for all $j_1 < j_2$, preserving the relative ordering of layer correspondences. Crucially, this formulation permits draft layer skipping, meaning that some draft layers may remain unmapped while others can be mapped to multiple verifier layers. This flexibility accommodates the common scenario where draft and verifier models have different depths and architectural characteristics. The problem is solved using dynamic programming where each state $(i,j)$ represents the minimum cost of aligning draft layers up to $i$ with verifier layers up to $j$, considering three transition options: diagonal matching (one-to-one layer correspondence), horizontal repetition (mapping the same draft layer to consecutive verifier layers), and vertical skipping (advancing through draft layers without correspondence).
Pseudo code to solve layer mapping problem is given in Algo \ref{algo:modified_dtw}.

\begin{algorithm}
\label{algo:modified_dtw}
\caption{Monotonic Dynamic Time Warping for Layer Mapping}
\begin{algorithmic}[1]
\Procedure{MonotonicDTW}{$S \in \mathbb{R}^{m \times n}$}
    \State \textbf{Input:} Score matrix $S$ where $S[i,j]$ is similarity between draft layer $i$ and verifier layer $j$
    \State \textbf{Output:} Mapping from draft layers to verifier layers
    
    \State Convert to distance matrix: $D = -S$ \Comment{For maximization}
    \State Initialize DP table: $dp[0,0] = 0$, all other entries $= \infty$
    \State Initialize backtrack table for path reconstruction
    
    \For{$i = 1$ to $m$} \Comment{For each draft layer}
        \For{$j = 1$ to $n$} \Comment{For each verifier layer}
            \State $cost = D[i-1, j-1]$
            
            \State \textbf{// Consider three alignment options:}
            \State $diagonal = dp[i-1, j-1] + cost$ \Comment{Match current layers}
            \State $left = dp[i, j-1] + cost$ \Comment{Repeat draft layer}
            \State $above = \min_{k=0}^{i} dp[k, j-1] + cost$ \Comment{Skip draft layers}
            
            \State $dp[i,j] = \min(diagonal, left, above)$
            \State Record best choice in backtrack table
        \EndFor
    \EndFor
    
    \State Find minimum cost in last column: $\min_{i} dp[i, n]$
    \State Backtrack from optimal endpoint to reconstruct path
    \State Convert path to layer mapping
    
    \State \Return layer mapping, total score
\EndProcedure
\end{algorithmic}
\end{algorithm}

\section{Stepwise cumulative perplexity}
\begin{figure}
  \centering
    \includegraphics[width=0.45\linewidth]{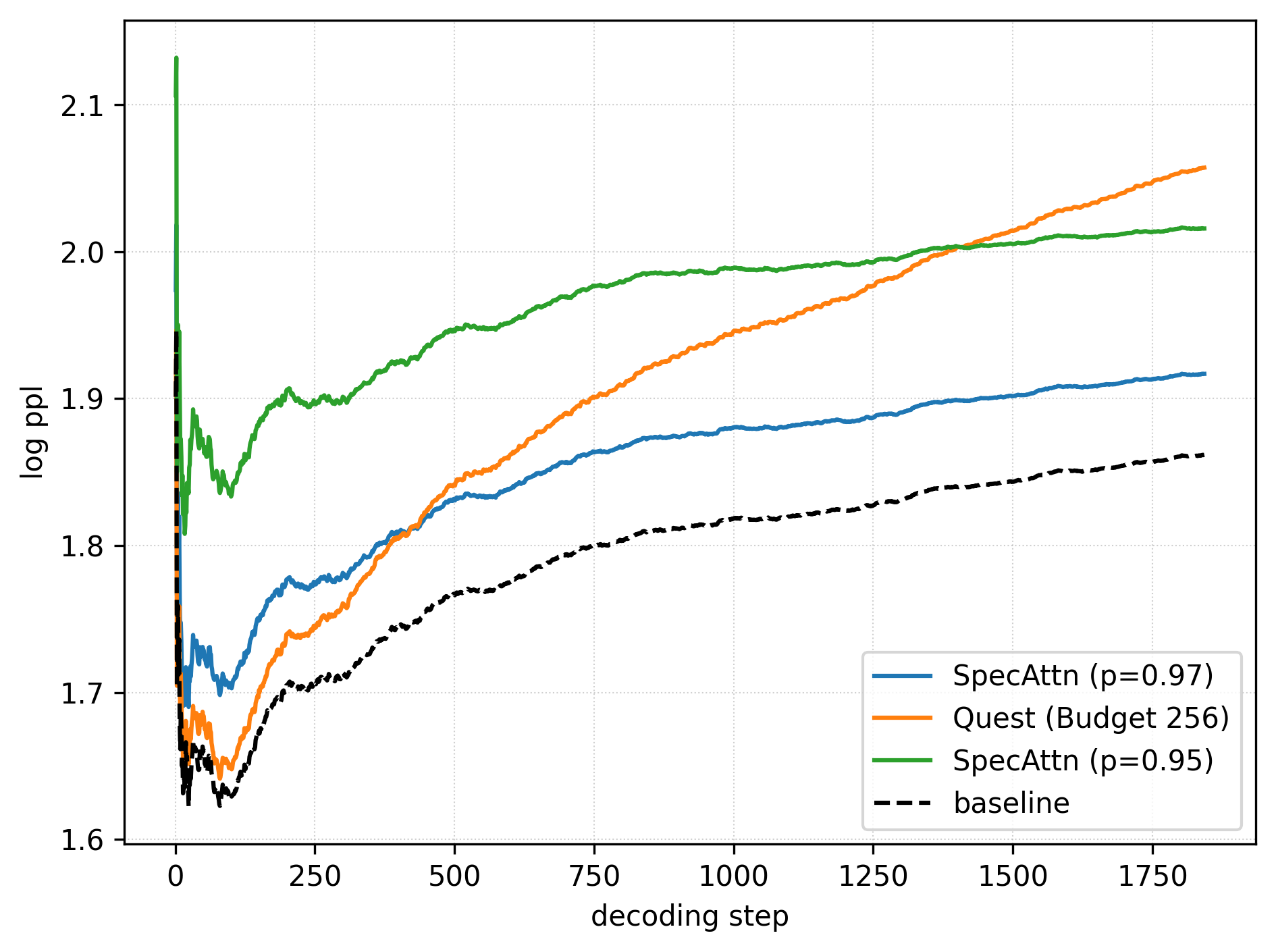}
  \caption{Perplexity (lower is better) comparison across different sparse attention methods. Here Baseline refers to the vanilla full attention decoding (StreamingLLM is omitted due to relatively high perplexity).}
  \label{fig:perplexity_comparison}
\end{figure}

Fig \ref{fig:perplexity_comparison} shows trends of log cumulative perplexity while decoding (on PG-19 dataset \cite{raecompressive2019}). Notice how Quest diverges faster than SpecAttn as the decoding step increases.

\end{document}